\documentclass{article}

     \usepackage[final]{neurips_2019}

\usepackage[utf8]{inputenc} 
\usepackage[T1]{fontenc}    
\usepackage{hyperref}       
\usepackage{url}            
\usepackage{booktabs}       
\usepackage{amsfonts}       
\usepackage{nicefrac}       
\usepackage{microtype}      
\usepackage{svg}
\usepackage[ruled,vlined]{algorithm2e}
\usepackage{amssymb,amsmath,amsthm}

\def\R{\mathbb{R}}
\def\supp{\mathsf{support}}
\def\X{\mathcal{X}}
\def\P{\mathsf{P}}
\def\on{\mathsf{on}}
\def\off{\mathsf{off}}
\def\dc{\mathsf{dc}}

\title{Finding Input Characterizations for Output Properties in ReLU Neural Networks}

\author{%
  Saket Dingliwal$^\star$ \\
  \texttt{sdingliw@cs.cmu.edu} \\
   \And
   Divyansh Pareek$^\star$ \\
   \texttt{dpareek@cs.cmu.edu} \\
   \And
   Jatin Arora$^\star$ \\
  \texttt{jatina@cs.cmu.edu} \\
}

\begin{document}
\newtheorem{lemma}{Lemma}

\maketitle

\begin{abstract}
Deep Neural Networks (DNNs) have emerged as a powerful mechanism and are being increasingly deployed in real-world safety critical domains. Despite the widespread success, their complex architecture makes proving any formal guarantees about them difficult \cite{ribeiro2016should}. Identifying how logical notions of high level correctness relate to the complex low-level network architecture is a significant challenge \cite{DBLP:journals/corr/abs-1907-10662}. In this project, we extend the ideas presented in  \cite{gopinath2019finding} and introduce a way to bridge the gap between the architecture and the high level specifications. Our key insight is that instead of directly proving the safety properties that are required, we first prove properties that relate closely to the structure of the neural net and use them to reason about the safety properties.
We build theoretical foundations for our approach, and empirically evaluate the performance through various experiments, achieving promising results than the existing approach by identifying a larger region of input space that guarantees a certain property on the output.
\end{abstract}

\section{Background}
\subsection{Notation}
We first setup notation for a ReLU neural network. Let $L$ be the number of hidden + output layers. $l \in \{0,1,\cdots L\} = [0,L]$ denote a layer in the network. $l = 0$ denotes the input layer, $l \in [1, L-1]$ denote the hidden layers and $l = L$ denotes the output layer. ReLU activation is applied only on the neurons in the hidden layers. $N_l$ denotes the number of neurons in layer $l$. Tuple $(l,i)$ refers to a neuron of the network in layer $l \in [1,L]$ and $i \in [0, N_l)$. $X \in \R^{N_0}$ is the input of the network and $Y = F(X) \in \R^{N_L}$ is the output -- interpreted as logits. Let $A_{(l,i)}(X)$ denote the feed-in (before applying ReLU) to the neuron numbered $i$ in layer $l$ on input $X$ to the network. \\
An activation pattern $\sigma$ specifies an activation status$\{\mathsf{on} /\mathsf{off}\}$ for a \textit{subset} of the hidden neurons (since the activation ReLU is applied only on hidden neurons). The neurons that are not specified are assumed to be ``don't cares". So equivalently $\sigma$ specifies activation status $\{\mathsf{on} /\mathsf{off}/ \dc\}$ for \textit{all} hidden neurons. A ReLU neuron is considered off when it's output is zero and on when it's output is positive. For a pattern $\sigma$, let $\sigma(l,i) = \on / \off/ \dc$ denote that $\sigma$ constraints the neuron $(l,i)$ to be on / off / dc respectively.
A pattern $\sigma$ is said to be $\mathit{complete}$ if $\forall (l,i), \sigma(l,i) \in \{\on, \off\}$. For any input $X_0$, let $\sigma_{X_0}$ denote the activation pattern dictated by the forward pass of $X_0$. Note that $\sigma_{X_0}$ is complete. \\
Let $\sigma_1 \preceq \sigma_2$ be a relation which holds when $\sigma_1$ is a sub-pattern of $\sigma_2$, which means that if $S_1$ denotes the set of neurons constrained by $\sigma_1$ and $S_2$ by $\sigma_2$, then $S_1 \subseteq S_2$ and $\forall (l,i) \in S_1, \sigma_1(l,i) = \sigma_2(l,i)$ (ie, the activation pattern of all neurons in $S_1$ agrees with that in $S_2$). \\
We define $\mathsf{support}(\sigma) = \{X \in \mathcal{X}: \sigma \preceq \sigma_X \}$. The size of this set is a measure of the number of inputs that follow the pattern. Let
$$\sigma(X) := \bigwedge\limits_{\sigma(l,i) = \on}  A_{(l,i)}(X) > 0\ \wedge \bigwedge\limits_{\sigma(l,i) = \off} A_{(l,i)}(X) \leq 0  $$
We say that a logical formula is convex if the set of inputs for which it is $\mathsf{true}$ is convex. We refer to the output vector of the neural net as $Y$ or $F(X)$. An element of this vector will be referred to as $Y_i$. For a given property $\mathsf{P}(Y)$, the aim is to identify convex sets in $\mathcal{X}$ such that $X \in C \Rightarrow \mathsf{P}(F(X))$. Since the property is also a logical formula on Y, the notion of convex properties is meaningful.\\
We say that $\sigma$ satisfies $P$, if $P(F(X))$ holds for all $X \in \mathsf{support}(\sigma)$. Otherwise, $\sigma$ invalidates $P$. $\sigma$ is minimal wrt a property $P$, if it satisfies $P$ and unconstraining any neuron in the pattern invalidates $P$. Minimality helps in getting rid of unnecessary constraints, and ensuring that more inputs can satisfy the property, by covering a larger region in the input space. 

\subsection{Theory}
Consider a complete activation pattern $\sigma$. It is easy to see that
$$X \in \supp(\sigma) \iff \sigma(X)$$
We summarize some of the results that were proven in \cite{gopinath2019finding} for all complete patterns $\sigma$.
$\supp(\sigma)$ is a convex set, equivalently, $\sigma(X)$ is a convex formula. Moreover, $\supp(\sigma)$ is simply an intersection of halfspaces. This is because $\sigma(X)$ is expressed as a logical and of various clauses, and each clause is of the type $A_{(l,i)}(X) \leq 0$ or $A_{(l,i)}(X) > 0$. And since each $A_{(l,i)}(X)$ is affine in $X$, the set becomes an intersection of halfspaces. Note that for the $\on$ constraints the equality on the halfspace is not included.
Next $ \forall l, \forall X \in \supp(\sigma), A_{(l,.)}(X) = W^\sigma_{(l,.)} \cdot X + b^\sigma_{(l,.)} $ where $W^\sigma_{(l,.)} \in \R^{N_l \times N_0} \text{, } b^\sigma_{(l,.)} \in \R^{N_l}$. This says that each intermediate neuron is an affine function of the input. Not just that, the output is an affine function of the input inside a complete $\sigma$. That is, there exist $W \in \R^{N_L \times N_0}, b \in \R^{N_L}$ such that $ X \in \sigma(X) \Rightarrow Y = W\cdot X + b$. We assume a function $\mathsf{getWeights}$ that takes an activation pattern $\sigma$ and returns $W, b$ for it. 
Also for any convex property $\mathsf{P}$,  $\sigma(X) \wedge \mathsf{P}(W\cdot X + b)$ is a convex region of space in $\mathcal{X}$ for which $\mathsf{P}(F(X))$ holds.

\subsection{Baseline}
For this discussion, we assume an implementation of the function $\mathsf{FindMinimal}(\sigma,\ P)$ which returns a minimal $\sigma'$ such that $\sigma' \preceq \sigma$ and $\sigma'$ satisfies $P$, based on the pre-condition that $\sigma$ satisfies $P$. We also assume the function $\mathsf{checkSAT}$ that takes an activation pattern and a property and checks if $\sigma \Rightarrow P$. The implementation details of these can be found in the Experimental section. \\
The paper uses the following algorithm to find activation patterns that satisfy $P$:

\begin{algorithm}[H]
\SetAlgoLined
 \SetKwInOut{Input}{input}\SetKwInOut{Output}{output}%
 \Input{
    Property $P$ and input $X_0$ such that $P(F(X_0))$ holds
 }
 \Output{
 Formula $Q$ on $\mathcal{X}: Q(X) \Rightarrow P(F(X))$
 }
 $\sigma \leftarrow \sigma_{X_0}$\;
 \eIf{$\mathsf{checkSAT} (\sigma, P)$}{return $\mathsf{FindMinimal}(\sigma,\ P)$}{ $W, b = \mathsf{getWeights(\sigma)}$\; return $\sigma(X) \wedge P(W.X + b)$\;}
 \caption{getConvexRegion($P, X_0$)}
\end{algorithm}

\section{Our Approach}
Consider a point $X_0$ that satisfies $P(Y) = Y_0 > Y_1$ and has the activation pattern $\sigma$. Let $W, b = \mathsf{getWeights}(\sigma)$. The method $\mathsf{getConvexRegion}$ will check if $\sigma \Rightarrow P(Y)$. If this check fails then the method returns $\sigma(X) \wedge P(W\cdot X + b)$. Therefore, in this case no minimisation of $\sigma$ takes place. Note that the property $P$ is high level and does not consider the architecture of the neural network. There is no general intuition that the activation pattern corresponds to the property that we are trying to check. Hence, it is expected that the method majorly returns $\sigma(X) \wedge P(W\cdot X + b)$. This observation is backed by our experiments, where this procedure enters the $\mathsf{else}$ branch frequently. \\
We propose that instead of finding the minimal activation pattern such that it implies $P(Y)$, we should find the minimal activation pattern that implies another property $\mathsf{I}(X, Y)$. The condition on $\mathsf{I}(X, Y)$ being that using $\mathsf{I}$, we should be able to represent the property $P$. As a trivial example, consider $\mathsf{I}(X, Y) := \big(Y_0 - Y_1 = (W_0 \cdot X + b_0 - W_1\cdot X - b_1) \big)$. We know that $\sigma \Rightarrow \mathsf{I}(X, Y)$ (because $Y = W \cdot X + b$ in $\sigma$). Therefore, the algorithm $\mathsf{getConvexRegion}$ at ($\sigma, \mathsf{I}$) will never go to the $\mathsf{else}$ branch. The algorithm then returns $\sigma' = $ $\mathsf{FindMinimal} (\sigma, \mathsf{I})$. \\
Now using the fact that $\sigma' \Rightarrow \mathsf{I}$, we can express a convex formula that implies $P$ as follows:  $$\mathcal{G}(X) := \sigma'(X) \wedge (W_0 \cdot X + b_0 - W_1\cdot X - b_1 > 0)$$ 
Note that $\mathcal{G}(X) \Rightarrow P(F(X))$.\\
This approach is similar to the concept of interpolants introduced in \cite{McMillan2003InterpolationAS}. The idea there is that in order to prove $A \Rightarrow B$, we prove $A \Rightarrow C$ and $C \Rightarrow B$. This has been immensely successful in the context of model checking and verification of safety properties\cite{Baier:2008:PMC:1373322}. The benefit of this approach in this context is that we can leverage the low level architecture of the neural net to construct such $\mathsf{I}$. Doing this is a relatively easier task because we can employ existing research done for analysing neural network behaviour.  In this report we explore the choice of one such $\mathsf{I}$ based on intuitions from \cite{DBLP:journals/corr/abs-1803-03635}.

Let $dW = W_0 - W_1$ and $db = b_0 - b_1$. Consider $\mathsf{I}(X, Y) := \big(Y_0 - Y_1 > dW \cdot X + db - \epsilon \big)$. Note that for this $\mathsf{I}$ as well $\sigma(X) \Rightarrow \mathsf{I}(X, Y)$ for any $\epsilon > 0$. Therefore, $\mathsf{getConvexRegion}(\mathsf{I}, X_0)$ will return  $\sigma' = $ $\mathsf{FindMinimal} (\sigma, \mathsf{I})$. Now we can express the convex formula as $\sigma'(X) \wedge (dW \cdot X + db \geq \epsilon)$, which implies $Y_0 > Y_1$. We will now explain our intuition behind the choice of this particular $\mathsf{I}$. 

We know that for $X \in \mathsf{\sigma}$, $Y_0 - Y_1 = dW\cdot X + db$. From the lottery ticket hypothesis \cite{DBLP:journals/corr/abs-1803-03635}, it is clear that there are neurons in the network that have a very small effect on $Y$. Therefore, those neurons will have a small effect on $Y_0 - Y_1$. We want to quantify this "small" change by introducing an $\epsilon$ relaxation. It might be possible to turn those neurons from $\on/\off$  to $\dc$ while remaining within the $\epsilon$ approximation of $Y_0 - Y_1$. Hence, it is expected that $\mathsf{FindMinimal}(\sigma, \mathsf{I})$ will be able to find smaller activation patterns. Note that this does not guarantee that we will find better convex regions. If we keep $\epsilon$ too large, we will find $\sigma'$ with a large $\mathsf{support}$. But the final convex region that implies $P$ is $\sigma'(X) \wedge (dW \cdot X + db \geq \epsilon)$. Although increasing $\epsilon$ has increased the size of the set $\sigma'(X)$, it decreases the size of the set represented by $(dW \cdot X + db \geq \epsilon)$. On the other hand, if we keep $\epsilon$ too small, we won't get considerable relaxation in the activation pattern $\sigma'$. We will explore the choice of $\epsilon$ in the experimental section, and illustrate this trade-off on a range of $\epsilon$ values. 

\subsection{Illustration}
In this section we take a very small ReLU neural network and illustrate some points of the theory. The input $X \in \R^2$ and the output $F(X) \in \R^2$. Staying in $\R^2$ makes visualization easier. There is one hidden layer with four neurons. We fix the weights and biases of this network. Let's focus on the property $\P(Y) = Y_0 > Y_1$.

\begin{figure}[!htb]
\minipage{0.32\textwidth}
  \includegraphics[width=\linewidth]{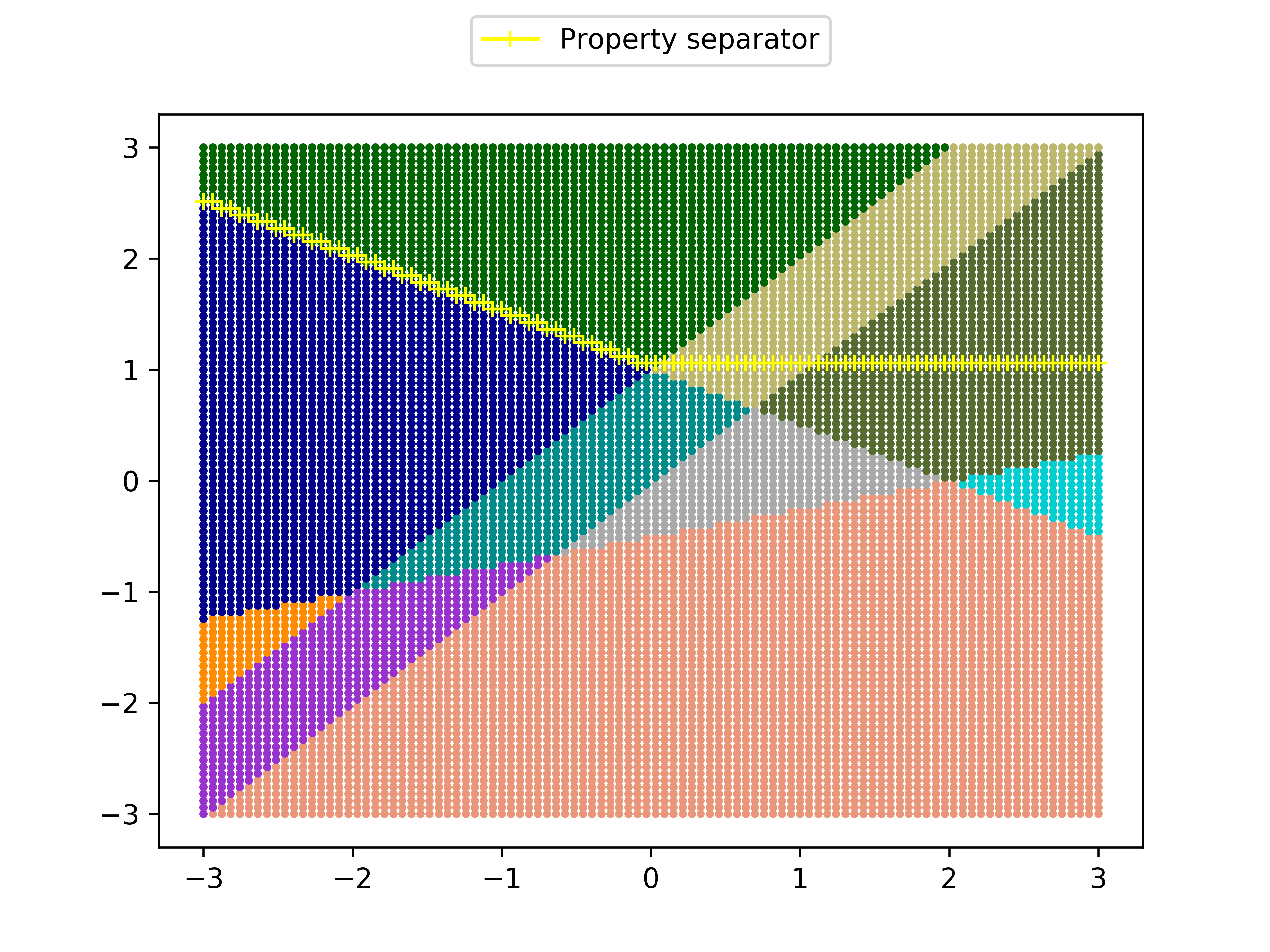}
  \caption{Honeycomb structure of Activation patterns}\label{fig:s1} 
\endminipage\hfill
\minipage{0.32\textwidth}
  \includegraphics[width=\linewidth]{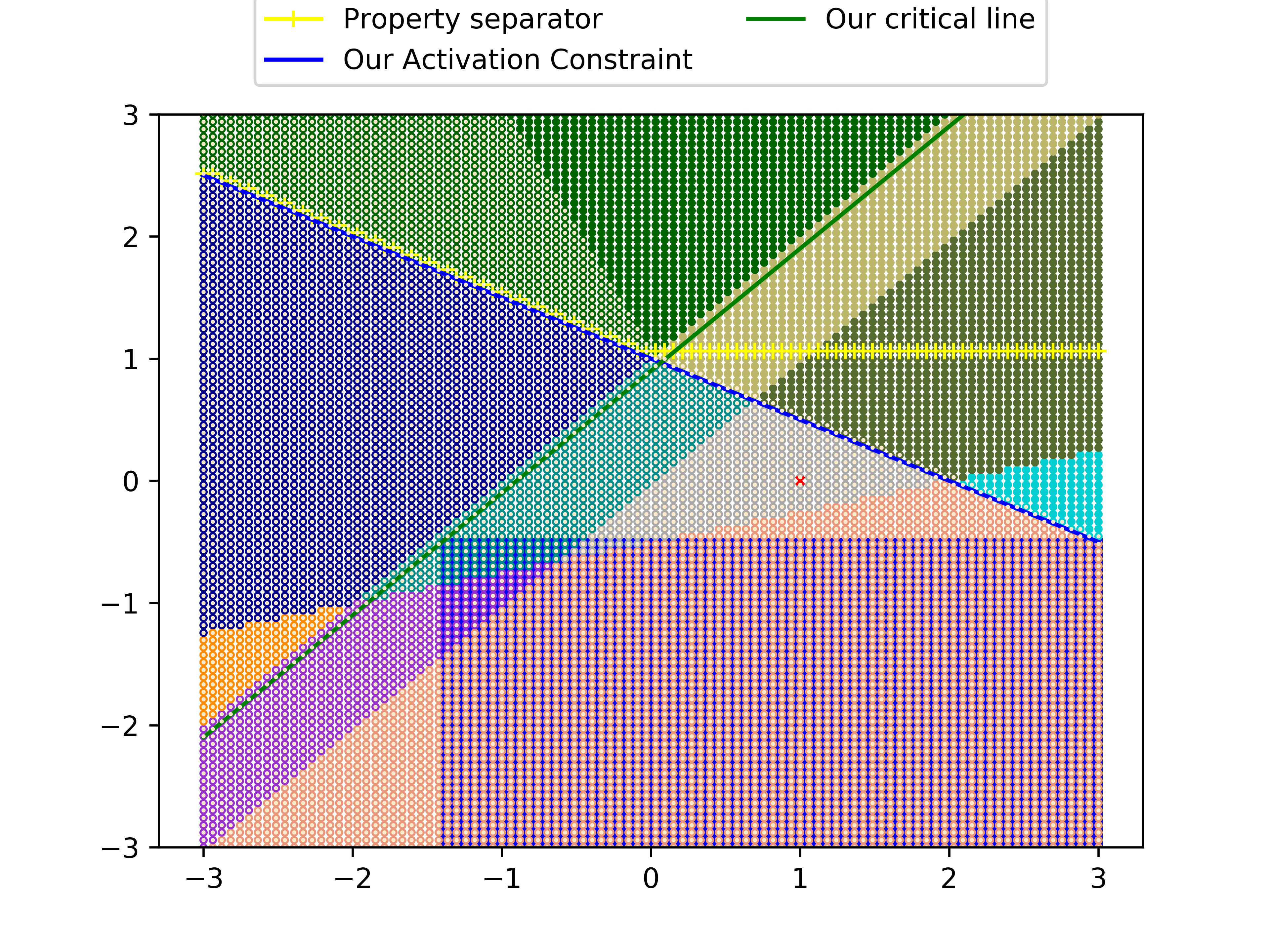}
  \caption{Our approach with $ \, \, \, \epsilon = 0.1 $}\label{fig:s2}
\endminipage\hfill
\minipage{0.32\textwidth}%
  \includegraphics[width=\linewidth]{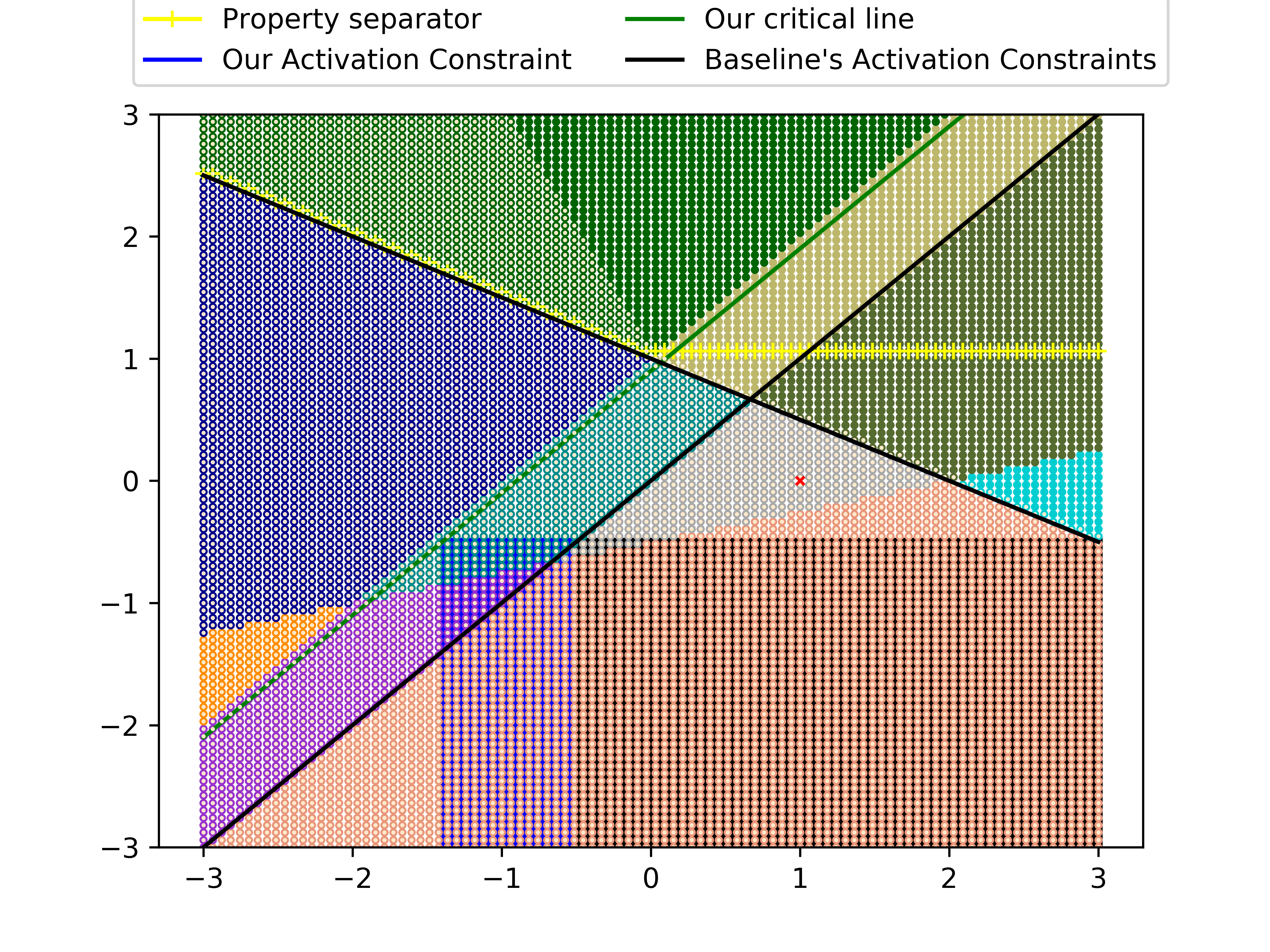}
  \caption{Baseline approach overlayed on our approach}\label{fig:s3}
\endminipage
\end{figure}

First we illustrate the ``honeycomb" structure of the various activation patterns. Each point $X \in \R^2$ lies in its activation pattern $\sigma_X$. Note that multiple $X$'s can have the same $\sigma_X$. There are a total of $2^{H}$ such patterns possible where $H$ is the number of hidden neurons. $H = \sum_{l=1}^{L-1} N_l = 4$ in this case. Of course, not all of them will be realizable because the weights of the network constraint what all activation patterns have a non-empty support. Because each activation pattern is an intersection of halfspaces, the entire structure of activation patterns overlayed on the input space ($\R^2$ here) forms a ``honeycomb"-like structure, shown in figure \ref{fig:s1}. 
There will be a separator in this space of the two classes that we predict. That separator will pass through some activation patterns. Inside an activation pattern the separator will be linear. This is because inside an activation pattern $Y_0$ and $Y_1$ are affine functions of $X$ so that the separator defined by $Y_0 = Y_1$ is linear. So the class separator is overall piecewise linear in the space $\X$. The yellow line in figure \ref{fig:s1} plots this separator of the original property, ie, the line at which $\P(Y)$ goes from True to False. 

We fix the starting point $X_0 = (1,0)^T \in \R^2$ to be the same for both the approaches (marked red cross in figures \ref{fig:s2} and \ref{fig:s3}). The activation pattern $\sigma_{X_0}$ has all 4 hidden neurons fixed. Our approach is able to relax this to $\sigma'_{our}$ that has only 1 neuron constrained (and other 3 are ``don't cares"). We then take the intersection of this region with the ``critical line" defined by $dW \cdot X + db \geq \epsilon$. This is illustrated in the figure \ref{fig:s2}. The blue box is the axis-aligned under approximation box of this region, which is a metric to compare regions of space (explained in the experiments section).

The baseline approach is able to relax 2 neurons out of the initial four to ``don't cares". So $\sigma'_{base}$ has two neurons constrained, shown in figure \ref{fig:s3}. Its axis-aligned under approximation box is shown in black.
We can see that our approach captures a bigger region of the input space. 

\section{Experiments}
\subsection{Implementation}
We implemented all approaches from scratch. We used Marabou \cite{katz2019marabou} as the decision procedure in the $\mathsf{checkSAT}$ calls. It takes the configuration of the neural network, the constraints on input, output \& internal variables as it's input and returns a point in the input space that simultaneously satisfies the constraints. Since we want to check $ \sigma \Rightarrow P$, we call Marabou with $\sigma \wedge \neg P$. An $UNSAT$ result of this call is equivalent to $\mathsf{checkSAT} (\sigma, P)$ being true. Borrowing the idea from \cite{gopinath2019finding}, we implemented $\mathsf{FindMinimal}$ using a greedy algorithm where starting from the last layer, we try to relax layers one by one. Once we reach a layer that cannot be relaxed as a whole, we try and relax neurons in this layer individually. For our experimentation with the choice of $\epsilon$, we define a hyper-parameter $\mathsf{logit-factor}$. We then choose $\epsilon$ to be $\mathsf{logit-factor}$ times the difference between logits of $X_0$ obtained through forward pass on the network. Such a choice makes $\epsilon$ robust to scale of the weights of the neural network as well as the input features. We used cvxpy \cite{cvxpy} to compute under approximation boxes ($\mathsf{UA-boxes}$) to evaluate the identified region in the input space. $\mathsf{UA-boxes}$ represent the axis aligned box that can fit inside the region given by the halfspaces in the input space. \cite{gopinath2019finding} used linear programming solver to compute under approximation boxes by maximising the objective of sum of ranges for each input feature. However, we changed the objective to maximize the volume in the feature and since the optimization problem then becomes non-linear, we used cvxpy. The code for the implementation is available. \footnote{\hyperref[https://github.com/typerSniper/NNInfer/tree/master/Code]{https://github.com/typerSniper/NNInfer/tree/master/Code}}

\subsection{Evaluation}
We designed and implemented various experiments to show effectiveness of our approach over the existing technique in the area. We first explain about the available datasets, followed by the metrics used for evaluation and the experimental setting. We then describe the experiments performed to motivate and evaluate different choices of the $\epsilon$. We conclude the section with the results of our algorithm in comparison to the baseline.

It is crucial to note that our approach can be used to verify any neural network with ReLU activations trained on any available classification dataset. Therefore, our choice of the dataset is guided by computational costs and interpretability. The data is used for predicting diabetes using 9 input features comprising of skin thickness, glucose level, blood pressure etc. It consist of 768 observations. We trained a neural network consisting of 2 hidden layers with 12 and 10 neurons respectively and ReLU activations. This network is able to achieve 81\% accuracy on using a train-test (85-15) split. We also evaluated our approach on standard MNIST dataset \cite{lecun1998mnist} using a deep ReLU neural network with 10 hidden layers consisting of 10 neurons each. Given the limitation of compute resources required to verify such a large network, we leave more extensive experimentation on this dataset as part of future work.

The output of our algorithm is an intersection of halfspaces in $\X$ for which the property to be verified holds. Therefore it is difficult to directly measure and compare the performance of similar algorithms in this domain. \cite{gopinath2019finding} proposes two different metrics for evaluation : (1) support on training data, and (2) volume of $\mathsf{UA-boxes}$. It is important to note that along with the bigger regions in the input space, we are also interested in finding those regions which carry the maximum probability mass of underlying data distribution. This is estimated by support in training data which is calculated as the number of points in the training data that lie in the output region in the input space.

\begin{figure}
    \centering
    \includegraphics[width=\columnwidth]{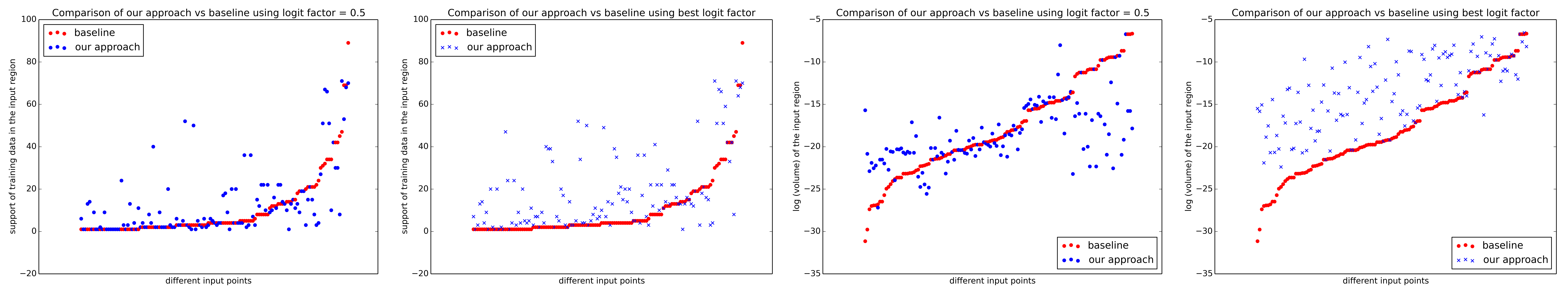}
    \caption{Comparison of metrics between baseline and our approach}
    \label{fig:overall}
\end{figure}
\begin{figure}
    \centering
    \includegraphics[width=0.6\columnwidth]{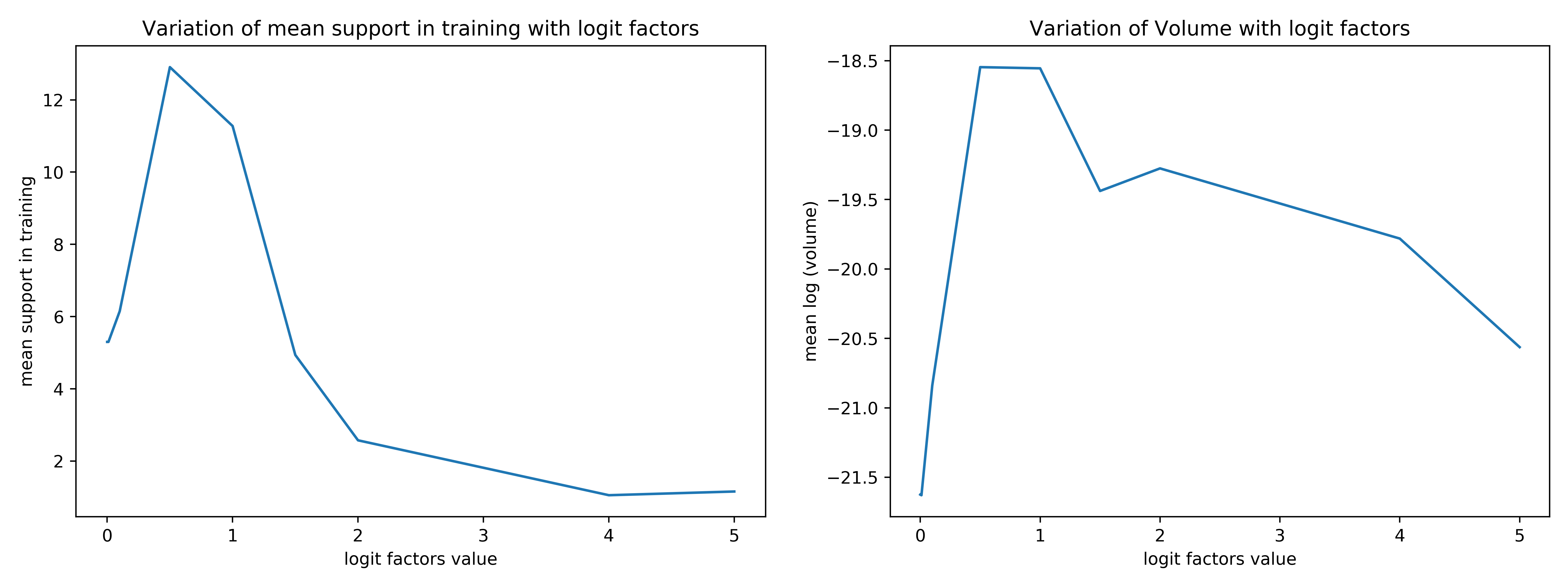}
    \caption{Performance of our approach for different values of $\epsilon \propto \mathsf{logit-factor}$}
    \label{fig:eps-exp}
\end{figure}

For the diabetes dataset, out of $2^{22}$ possible activation patterns in the network, only 127 of them are realized by the training set. 
This small number gives us freedom to run our algorithm on different $\epsilon$ values as well as the baseline on one of the input from each activation pattern. For each activation pattern, we try to find the maximum input region for the property that the predicted class of the given input is the chosen class for that region. Figure \ref{fig:overall} summarizes our experiments for comparison of our method with the baseline. It consists of four scatter plots where each red point corresponds to metric value for the baseline on one of the 127 initial input points while blue is for our approach. The first two plots correspond to the support in training data metric and the remaining two are for the log-volume for the $\mathsf{UA-boxes}$. The first plot for each metric shows the performance of our approach on a fixed hyper-parameter while the second one demonstrates the effectiveness of our approach if we chose the best hyper-parameter for each input. We see that for both the metrics, the majority of the blue points lie above their corresponding red, indicating that we perform significantly better. 

We validate our hypothesis about trends in choice of $\epsilon$. As mentioned earlier, a large $\epsilon$ allows more relaxation of the activation pattern but reduces the intersection region with the additional constraint of $dW \cdot X + db \geq \epsilon$. We observe this trend for all the inputs of the diabetes dataset we tried. We plot the average support and volume over all the points for different $\mathsf{logit-factor}$ in \ref{fig:eps-exp}. Both the plots show a region of increase followed by a decrease with the best performance achieved at $\mathsf{logit-factor} = 0.5$ 

\vspace{-5pt}
\subsection{Results}
We see a remarkable improvement over the baseline in our experiments. For the diabetes dataset, in 50 out of 127 initial points we begin with, we are able to achieve more training data support of the identified region, while an equal train support in 41 of the remaining points. This is achieved at $\mathsf{logit-factor} = 0.5$. This indicates that we are able to identify bigger regions and hence can reason about the properties for larger number of points from the data distribution. We also verify experimentally that in 53 initial points, the baseline is even not able to call $\mathsf{FindMininimal}$ as the property fails to hold true in the initial activation pattern. Since we guarantee call to $\mathsf{FindMininimal}$ in our approach, we demonstrate that we are more robust to the choice of the initial activation pattern. Finally, we supplement our findings about our approach by running it on larger network for MNIST. Given limited computational resource and expensive $\mathsf{checkSAT}$ calls, we compare ourselves with the baseline on two different inputs. We are able to successfully improve the train support from 343 to 397 in one of those inputs. This further demonstrates the effectiveness of our approach across different networks of variable sizes. 


\section{Future Work}
Building the theory and doing the experiments in this project has opened up a bunch of different directions that can be followed further. One is to design and experiment on properties other than class dominance (ie, $Y_0 > Y_1$). Secondly, it is clear that the choice of the initial point / activation pattern plays a crucial role in how big a region we can output. It would be worthy to try to characterize this mathematically. Further, \cite{gopinath2019finding} introduces layer invariants, which is another way to do the minimalization of activation pattern. It would be notable to experiment with relaxed properties in that setting. Finally, a major advancement could be made if we could reason about operations other than ReLU in a network. If we can build the theory to handle convolution and pooling operations, we could apply this technique to modern networks and verify properties on deep networks for large-scale datasets.

\bibliography{ref}
\bibliographystyle{plain}

\end{document}